# Knowledge distillation with error-correcting transfer learning for wind power prediction

Hao Chen

*Department of Technology and Safety, UiT The Arctic University of Norway, Tromsø 9019, Norway*

*Institute of Industrial Science, The University of Tokyo, Tokyo  153-8505, Japan*

## Abstract

Wind power prediction, especially for turbines, is vital for the operation, controllability, and economy of electricity companies. Hybrid methodologies combining advanced data science with weather forecasting have been incrementally applied to the predictions. Nevertheless, individually modeling massive turbines from scratch and downscaling weather forecasts to turbine size are neither easy nor economical. Aiming at it, this paper proposes a novel framework with mathematical underpinnings for turbine power prediction. This framework is the first time to incorporate knowledge distillation into energy forecasting, enabling accurate and economical constructions of turbine models by learning knowledge from the well-established park model. Besides, park-scale weather forecasts non-explicitly are mapped to turbines by transfer learning of predicted power errors, achieving model correction for better performance. The proposed framework is deployed on five turbines featuring various terrains in an Arctic wind park, the results are evaluated against the competitors of ablation investigation. The major findings reveal that the proposed framework, developed on favorable knowledge distillation and transfer learning parameters tuning, yields performance boosts from 3.3 % to 23.9 % over its competitors. This advantage also exists in terms of wind energy physics and computing efficiency, which are verified by the prediction quality rate and calculation time.

**Abbreviations**

| | |
|---|---|
| NWP | Numerical Weather Prediction |
| TL | Transfer learning |
| LSTM | Long Short-Term Memory |
| EDLSTM or ED | Encoder-Decoder Long Short-Term Memory neural networks |
| KD | Knowledge Distillation |
| MMD | Maximum Mean Discrepancy |
| MSE | Mean Square Error |
| Bi-LSTM | Bidirectional LSTM |
| RMSE | Root Mean Square Error |
| QR90 | Qualification Rate at the 90% threshold |
| GeoMean | Geometric mean |
| STD | Standard deviation |
| T#. | Wind turbines with different terrain – turbine number |

# 1. Introduction

The exploitation of renewable energy is a propitious approach to achieving carbon neutrality. Wind energy, as one of the foremost renewable energy sources, has gained increasing prominence in various countries for its abundant availability, technological maturity, and favorable financial support [1]. However, the volatility and randomness of the natural wind, especially when sited in complex terrain conditions [2], create tremendous uncertainty in wind power generation. The volatile electricity generated by wind turbines causes adverse effects on power quality, grid dispatch, and the stability and security of power system operation [3]. Therefore, as larger-scale wind power is integrated into the grid, there is a considerable necessity to further develop associated energy forecasting strategies [4] to reduce the impact of renewable on the grid and improve energy efficiency.

Wind power prediction can be categorized into physical, statistical, and hybrid approaches [5]. The first is appropriate for relatively long-term (days) forecasts through atmospheric physics modeling called. The second is suitable for relatively short-term (minutes or hours) predictions. However, the current prediction investigations, which usually combine the above two methods and can extrapolate the prediction-time scenario to realize higher accuracy, are attracting more interest [6].

## 1.1. Previous work

Extensive research has been conducted using Numerical Weather Prediction (NWP) downscaling [7], [8], signal decomposition [4], statistical regression [9], machine learning [5], [10], deep learning [11], and more for wind power prediction and has achieved satisfactory results in many wind park experiments.

The state-of-the-art of employing deep learning associated approaches to predict wind power is sufficiently summarized and reviewed in [11], [12], [13]. These studies are primarily devoted to the decomposition of wind energy-related sequences using different data processing techniques and developing advanced deep learning algorithms to recognize the in-depth features of these sequences. Since wind power is primarily driven by weather, adding information from suitable weather forecasts to a generation prediction mode can lead to an accuracy boost [14]. However, it is common to downscale mesoscale NWP (wind park-scale or larger) applying meteorological [8] or statistical methodologies [15], [16] or to input these data directly into the prediction model for automatic recognition [17], [18]. The former involves substantial physical assumptions and computational time whereas the latter loses differences between turbines in the same wind park.

As the availability of wind energy data and the accuracy of power prediction increases, so does the complexity of the forecasting models. Since different researchers have developed various forecasting models for the individual park, when faced with the problem of a new site, it is required to develop a fresh model from scratch.

Transfer learning (TL) is a machine learning paradigm that applies some kind of learning algorithms to derive information from one or more application scenarios and thereby contribute to enhancing the learning performance for the target scenarios [19]. There appear to be a few efforts in recent years to apply TL to wind power prediction. Hu Q et al. [20] tried to transfer wind speed information from a data-rich old operating wind park to a newly-built park and successfully used stacked denoising autoencoders to forecast the speed with each other data for four wind parks in the north of China. Qureshi AS et al. [21] speeded up and optimized the deep sparse autoencoder power prediction model learning by only training one park data and transferring the gained knowledge to others with sounds weight initializations. Yin H et al. [22] focused on multiple wind parks power forecasting with TL in Inner Mongolia and applied CNNs-LSTM to extract sequence features of source parks and deliver the features to build a good model for the target wind farm. These studies all directly and successfully transferred information from other wind farms to the intended one, but without accounting for the forecasting issue for turbines. Liu X et al. [23] considered turbines prediction by a proposed a deep and TL framework with the turbine Supervisory Control And Data Acquisition (SCADA) data, and by processing the homogeneity and heterogeneity among different wind turbines, they realized the ultra-short-term power prediction within one hour with high accuracy. However, the study is a purely statistical and short-term forecast; to achieve the long-term prediction, weather forecasts must be factored.

The forecasting error includes unlearned information, analyzed in [24], which can be extracted as a complementary for park power prediction models. A few studies consider the magnitude of prediction errors. Ding M et al. [25] used a gated recurrent unit neural network to correct the NWP wind speed error and applied the corrected speed to model the power curve with efficiency. Sun Z et al. [26] directly modeled the error sequence trend of preliminary prediction results and combined the predicted error to obtain the final predictions. However, most of these studies treated the prediction errors as separate time series, which are not always reliable because the error series may be very white-noisy and hard to be forecasted or the accuracy of their prediction decreases rapidly with the forecasting time step [27].

### 1.2. Contributions

As our previous work published in this ***journal*** *Data-augmented sequential deep learning for wind power* forecasting and its literature showed that the work returned to the physical process of wind power generation, the statistical characteristics of wind data, and the nature of deep learning to approach the prediction problem for the turbine in varying terrain conditions [17] and yielded a delicate forecasting structure promising multistep prediction.

However, there are still the following points where enhancements are possible according to our further investigations.

1. Deep learning-based models require multiple layers of uniquely designed neural network structures to realize the intrinsic features of the data [28]. And training large deep networks is very time-consuming and computationally intensive. Therefore, it is proposed that certain pre-training techniques, such as Knowledge Distillation (KD) [29], can be adopted to *distill* useful information, knowledge, from the large pre-trained, whole park, *teacher* model and *condense* it into smaller scale turbine prediction, *student*, models.

2. The NWP information in hybrid prediction models is usually for the whole wind park and does not precisely reflect the individual turbine's future meteorological condition. Besides, the wake effect (wind is strongly perturbed, decreased kinetic energy and added turbulence behind blades of a turbine), together with the turbulence induced by the micro-scale topography in complex terrain, thereby rendering further forecasting difficulties. Typically, single wind turbine meteorological modeling considers NWP results and simulates turbine wind conditions with Computational Fluid Dynamics (CFD) [30], [31]. This paper will bypass this complex approach based on multiple physical assumptions and indirectly integrate the meteorological information of turbines into the whole prediction model by data science.

3. Wind power forecasting is essentially reducible to a regression problem, so regression diagnostics in statistics, especially error analysis and correction, could be incorporated into the prediction model. Hence, this paper achieves the detection and forecasting of prediction errors through advanced deep learning approaches. Moreover, weather information is ingeniously embedded into the final prediction model by the errors correcting.

The presented paper fully addresses the above concerns by innovatively developing a framework for predicting wind turbine power with solid mathematical derivation. It is the first time, through KD, to deploy a large sequential deep learning forecasting model with multiple inputs and outputs for big data in wind parks on a fast-running small-scale turbine forecasting model; the framework fully exploits the prediction error value and ingeniously downscales the NWP information corresponding to wind parks non-explicitly to the turbine scale by transfer learning and primitive and inverse function transformation. The effectiveness and quickness of the framework are experimentally verified on wind turbines in different terrains. Furthermore, the framework has extensive applicability in other fields since it does not involve specific energy-physics and geographical factors.

The rest of this article is organized as follows. Section 2 elaborates on the deep learning techniques involved in the proposed wind power prediction framework. Section 3 describes case study experiment data and vividly deduces the mathematical principles underlying the framework. In section 4, the experimental procedure and evaluation are briefly stated. The experimental results for comprehensive verification of the framework are thoroughly discussed in section 5. Finally, the main findings and policy recommendations are exhibited in Section 6.

## 2. Methodology

This section elaborates on the peripheral techniques utilized in this paper, starting with Transfer Learning (TL), followed by Knowledge Distillation (KD), an important TL technique, and finally, prediction algorithms and their resultant error corrections.

### 2.1. Transfer learning

An essential challenge in employing machine learning applications in engineering is the weak displacement of models, i.e., existing models do not transfer well into new areas. Firstly, Many machine learning application applications are trained with small data: traditional learning algorithms tend to suffer from overfitting problems due to small data size [32]. Thus, the established models cannot be well extended to new scenarios. Secondly, strong robustness is required for machine learning models: classical algorithms assume that training and testing data are from the same statistical distribution [33], which are sometimes unrealistic in engineering. Finally, personalization and data security; in largescale practical applications, where datasets often belong to multiple owners and cannot be disclosed to each other for security reasons. Therefore, the learning model should extract the intrinsic of each dataset and transfer to new scenarios [34].

The emerging TL responds to this problem. It is a methodology that addresses data distributions in the source domain (where the model has been trained) and the target domain (where the trained model will be implemented), which are similar but not identical, to promote efficiency in machine learning [35]. The core in a sound TL is finding the similarity between the mentioned two scenarios to achieve a so-called adaptive learning. TL is categorized into two types with the based on the features in the two scenarios: homogeneous and heterogeneous [36]. The former is only considered because the data used are only wind sequence data and designed transfer process is elaborate. TL, especially deep TL, directly improves performance on different tasks by recognizing patterns directly on the original data and transferring the models to other original data. It enables the automagical extraction on more expressive features and meets the end-to-end demands of real-world applications.

In the present study, TL architecture is established with the development of an adaptive sparse deep learning network for univariate pattern recognition. The structure is shown in Fig. 1.

In fact, adaptive learning is effectively applied by identifying and reducing differences between the source and target domains through simple and executable transformations to transfer learned models from source domains to target ones. The Kullback Leibler Divergence ($D_{KL}$), also named relative entropy in (1), is mostly employed to identify similarities between two distributions $p(x)$ and $q(x)$. Practically, Practically, since $D_{KL}$ involves integration and logarithmic operations, a similar metric Maximum Mean Discrepancy (MMD), in (2), as an approximation of bivariate divergence is available in the [37].

$$D_{KL}(p(x) \parallel q(x)) = \int_{-\infty}^{\infty} p(x)\ln\frac{p(x)}{q(x)}dx \tag{1}$$

$$MMD(x, x_t) = sup\left(\frac{1}{N}\sum_{i=1}^{N} f(x^i) - \frac{1}{N_t}\sum_{j=1}^{N_t} f(x_t^j)\right) = \left[\frac{1}{N^2}\sum_{i,j=1}^{N} f(x^i, x^i) - \frac{2}{N \cdot N_t}\sum_{i,j=1}^{NN_t} f(x^i, x_t^j) + \frac{1}{N_t^2}\sum_{i,j=1}^{N_t} f(x_t^i, x_t^j)\right]^{\frac{1}{2}} \tag{2}$$

where $sup(.)$ is supremum; $x^i$ and $x_t{}^j$ are the $i$-th and $j$-th sample in the source and target domains, respectively; $f(.)$ donates feature mapping function in TL; $N$ and $N_t$ represents source and target domain size.

For deep learning network (extensively literature detailing DL [28]). The utilized sparse DL's (shorten as $D$1) loss function $L_{D_1}$ is expressed as follows:

$$L_{D1} = \frac{1}{n}\sum_{i=1}^{n}(X_i - \hat{X}_i)^2 + \beta\Omega_w + \delta\Omega_s \tag{3}$$

where $\frac{1}{n}\sum_{i=1}^{n}(X_i - \hat{X}_i)^2$ is the Mean Squared Error (MSE), typical loss function for regression task, $\Omega_w = \frac{1}{2}\sum(W_{i,j})^2$ donates the $L_2$ regularization term (handling over fitting) of NN weights, $\Omega_s = D_{KL}(p_i \parallel \hat{p}_i)$ represents the sparse regularization term that enforces constraints of sparsity in outputs from hidden layers to minimize the calculated data distribution $\hat{p}_i$ and inputs actual distribution $p_i$.

The well-learned model is normally can be directly transferred to process new datasets. Since the old and new datasets are merely similar but not identical, an adaptive MMD-based mechanism (an Optimizer named MMD comparator) is introduced into the model transfer to further improve the new model's ($D$2) accuracy. The loss function $L_{D_2}$ of $D$2:

$$L_{D2} = L_{D1} + \gamma\Omega_{MMD} \tag{4}$$

where $\Omega_{MMD} = \sum_{layers} MMD(X_i, X_i')$ is the divergence regularization term to minimize the gaps between the layer to layer of old and new models. So, $L_{D_2}$ optimized the transferred old model to fit new data better.

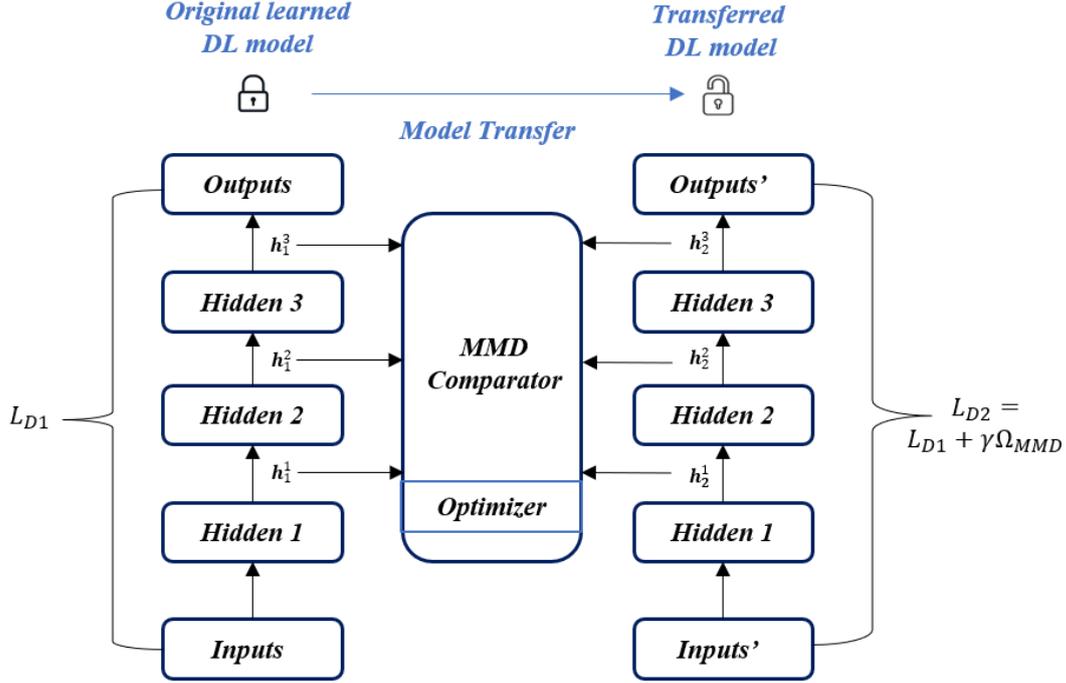

Fig. 1. The architecture of proposed TL with adaptive sparse deep learning

## 2.2. Knowledge Distillation

Distillation is a chemistry term referring to the extraction of components at different boiling temperatures. Typically, big and complex models that have excellent performance and generalization, but with enormous parameters are hard to be deployed, which is not available to small models. Inspired by pedagogy, in a similar problem, whether the knowledge acquired from big models could guide small ones, so that the latter, meanwhile with reduced parameters' number, have improved performance and generalization ability like the former.

KD [38], based on the "teacher-student" learning strategy, is a model compression approach, and its application is emerging in academia and industry due to its conciseness and effectiveness [39]. Specifically, it means relocating the pre-trained knowledge from teacher networks, typically large or ensemble, to student networks that are relatively small or simple, so that the student's learn the know-how from the teacher's network. The KD system generally involves three pivotal components: knowledge, distillation algorithm, and teacher-student architecture [39]. The KD with the teacher-student transfer structure is vividly shown in Fig 2.

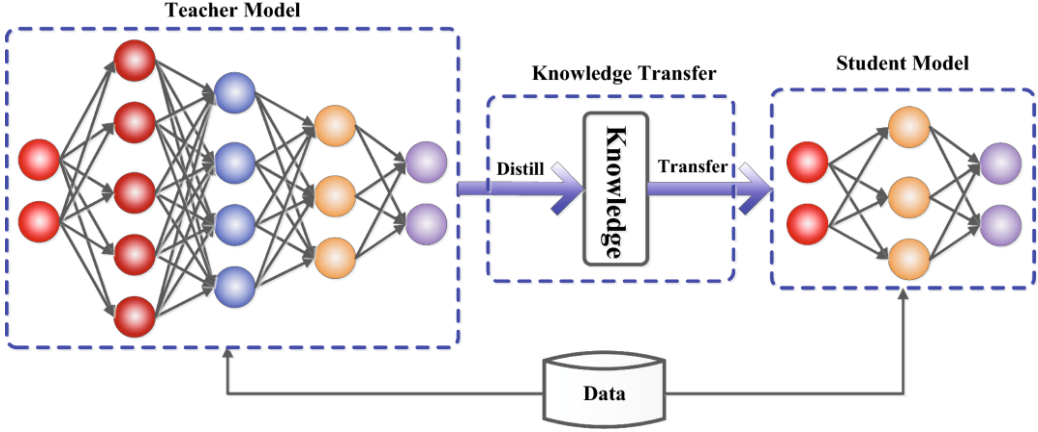

Fig. 2. The general structure of knowledge distillation.

While former KD studies have focused on classification tasks, few studies concentrated on regression missions, such as speech recognition [40] and objects localization [41], etc. There is a scarcity of investigation on KD approaches for time-series tasks., which are fully developed in the present study.

### 2.3. Proposed KD forecasting framework

In the present study, inspired by [42], [43], and [44], the KD networks architecture shown in Fig. 3, which is implemented to distill the wind park prediction model with big data to turbine prediction. The proposed KD forecasting framework is developed to address three challenges.

1. Large stacked LSTM-based encoder-decoder models are adopted to predict the wind park with big data while small Bidirectional LSTM (Bi-LSTM) models are applied to forecast small data turbines.
2. Cross-dataset knowledge transfer, model to model, in KD is achieved by developing a suitable operator that allows different teacher and student models, based on different inputs, in the KD structure to be compared in training.
3. The designed linear combination of *Loss* I and II in designing student model's loss function *Loss* KD realizes the optimization of the student model not only by training from its own data but also by further learning from the high-performance complex teacher model.

The operation of this KD architecture is as follows: firstly, the wind park big data is imported into the large teacher model, as the pre-trained, to learn the data depth features and generate the predicted park power. Simultaneously, the turbine data are also fed into the student model for preliminary learning and to produce preliminary turbine power calculations. Then the fine-tuning process with KD for the student model begins with the mission of minimizing its loss function *Loss* KD. In regression tasks with KD, the loss function can be generally expressed in (5)

$$L_{\text{reg}} = \frac{1}{n}\sum_{i=1}^{n} min(\|\hat{\mathbf{p}}_S - \mathbf{p}_S\|^2, \|\hat{\mathbf{p}}_S - \hat{\mathbf{p}}_T\|^2) \tag{5}$$

where $\hat{\mathbf{p}}_T$ donates output of the pre-trained teacher model, $\hat{\mathbf{p}}_S$ is outputs of primarily student model, and $\mathbf{p}_S$ is corresponding real values of inputs of the student model. Notably, since these park and turbine predicted power statistics are not identical, operators are necessary to scale them and to compare each other. The mentioned **p** values are defined as in Eq. (6) and calculated with the two operators.

$$\hat{\mathbf{p}}_T = \left\|\frac{P_p - \hat{P}_p}{P_p}\right\|, \hat{\mathbf{p}}_S = \left\|\frac{P_t - \hat{P}_t}{P_t}\right\|, \mathbf{p}_S = \mathbf{0} \tag{6}$$

Simplifying the function to a linear combination of two loss functions as in

$$L_{\text{reg}} = \frac{1}{n}\sum_{i=1}^{n} \alpha\|\mathbf{p}_S\|^2 + (1-\alpha)\|\hat{\mathbf{p}}_S - \hat{\mathbf{p}}_T\|^2 \tag{7}$$

To prevent overfitting and reduce model complexity, the KD will stop when the student model has a lower error than the teacher one. The formula (7) can be further simplified as

$$L_{\text{reg}} = \frac{1}{n}\sum_{i=1}^{n} \alpha\|\mathbf{p}_S\|^2 + (1-\alpha)L_{\text{compare}} \tag{8}$$

$$L_{\text{compare}} = \begin{cases} \|\hat{\mathbf{p}}_S - \hat{\mathbf{p}}_T\|^2, & \text{if } \|\hat{\mathbf{p}}_S\| > \|\hat{\mathbf{p}}_T\| \\ \mathbf{0}, & \text{otherwise} \end{cases} \tag{9}$$

where $P_p$ and $P_t$ are measured power of park and turbine, $\hat{P}_p$ and $\hat{P}_t$ and their predicted power by the teacher and student models, respectively. $0 < \alpha < 1$ represents the vital index named KD parameter that controls the balance between knowledge flow from teacher and student itself in the learning stage.

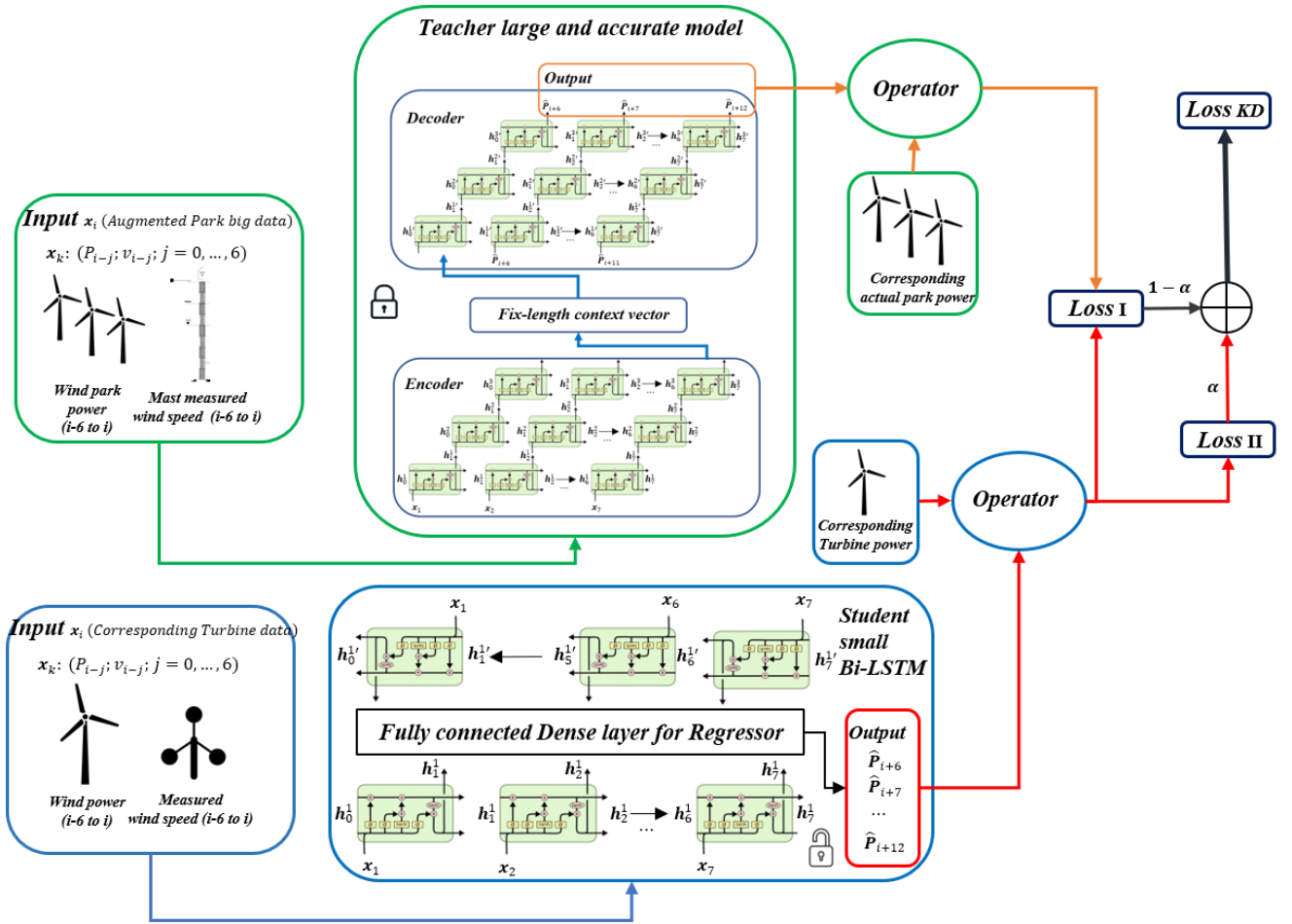

Fig. 3. The proposed KD structure for distilling knowledge from park to turbine.

## 2.4. Error correction for NWP

As described in Section1, NWP data are often on large scales and can often be downscaled from tens of kilometers to a few ones as applicable to wind parks. However, research has shown that a lower scaling is not always advisable in wind power prediction [45]. Also, it is difficult to accurately downscale NWP to wind turbines (tens of meters) because of the involved turbulence and sophisticated CDF assumptions. In engineering practice, the NWP wind data corresponding to the park are accessible. Meanwhile, the turbine data are analogous to those of the park. In view of the profound pattern mining and matching capabilities of deep learning and the neat structure of TL, park's NWP information can be refined and transferred to more accurate turbine predictions through the corresponding data-driven prediction error correction approach. The proposed NWP error correction approach is explained in detail in section 3.2.

**2.5. Modular for Forecasting algorithms**

The research employs three deep learning regression algorithms: a multilayer general deep neural network for univariate TL (error correction from park to turbine and integration of turbine training and testing), Bi-LSTM for student prediction of turbines model, and a developed large-scale EDLSTM for forecasting the wind park, respectively. The first two are well-established deep learning models that are well elaborated in [28]. And the superiority of EDLSTM, applied in the proposed KD structure for park prediction algorithms, compared to other machine learning benchmark models (Persistence, simple three-layer backpropagation Neural Networks, basic LSTM, bionic optimized neural networks constructed Adaboost ensemble learning) has been thoroughly analyzed in [17]. And the detailed description of competitors for the proposed prediction framework is described in section 5.2. Since modern deep learning models are modularly constructed, a brief summary of the used modules is presented in Table. 1.

Table 1. A brief summary of prediction modules' setup

| Forecasting model | Main parameters |
|---|---|
| Fully connected layer | Sigmoid activation function, and Mean Square Error (MSE) loss function |
| Deep neural network for univariate TL | The input, 3 hidden, and output layers are with 7, (15, 10, 15) and 7 neurons, respectively; sigmoid activation function, and MSE loss function. (Hidden layers neurons numbers are roughly found with a grid search with a unit of 5 from 5 to 30.) |
| LSTM unit | Sigmoid activation function, MSE loss function, and Adam algorithm optimizer. |
| Bi-LSTM | Fully connected layer for internal outputs and the regressor, Two-layer seven LSTM units with reverse directions A fully connected layer with 7 neurons as the external output layer. Dropout avoids overfitting. |
| EDLSTM | The encoder and decoder, linked with a fixed-length context layer, are with 3 layers of 7 connected LSTM units. Dropout and early stop to avoid overfitting. Batch size control for fast learning. |

# 3. Data preparation and prediction theory

## 3.1. Data preparation

Wind turbines can convert the kinetic energy from the wind into electricity. Theoretically, the wind power generation model is expressed in (10):

$$P = \frac{1}{2}\rho\pi R^2 C_p(\lambda,\beta)v^3 \tag{10}$$

where $P$ is the wind turbine power generation (W); $\rho$ is the air density (kg/m²); $R$ donates the rotor radius; $C_P$ means wind energy utilization efficiency (dependent on the tip speed ratio $\lambda$ and the blade pitch angle $\beta$); $v$ represents the wind speed (m/s). According to (1), the output of a wind turbine is mainly about the wind speed, air density, swept area, and blade conditions, all of which are strongly influenced by the local weather of the turbine.

Given that the present investigation is a follow-up study to the previous article [17], the data from the same wind park, a 54 MW with 18 Vestas V90 3 MW turbines with complex terrain in the Arctic, are maintained for comparability. The measured 2017 year data with a 10-mins temporal resolution of the whole wind park (also with 2.5 km spatial and 1 h temporal resolution NWP data) and 5 turbines with varying terrains (T1 Plateau, T2 Valley, T3 Lakeside, T4 Hilltop, T5 Seaside).

For park and turbine, the raw 10-mins measured data derived from the averaged interpolation approach are utilized since it is required to forecast its hours-ahead power. The details of the used data can be found in Section 2 of [17].

Due to the varying physical variables involved in the mentioned equations, the raw data should initially be scaled before learning with data standardization (adjustment by placing the mean at 0 and the standard deviation at 1) to facilitate the convergence speed of the loss function in learning [46].

### 3.2. Prediction theory

Mathematically, wind power prediction with deep learning can be simplified to a function construction problem. Where the variables can be linked by suitable function operators. This paper proposes the following functional approach to realize transferring large wind park power forecasting models to small turbine prediction and realize non-explicit NWP downscaling by error correction.

**Algorithm 1.**

1. Teacher (large) models for wind park $p$ and their error functions.

$$\begin{aligned}
\hat{P}_{i+n_p}^{I} &= f_p^{I}\left(P_{i-j_p}; v_{i-j_p}\right), \\
Error_p^{I} &= P_{i+n_p} - \hat{P}_{i+n_p}^{I}, \\
\hat{P}_{i+n_p}^{II} &= f_p^{II}\left(P_{i-j_p}; v_{i-j_p}; u_{i+n_p}\right), \\
Error_p^{II} &= P_{i+n_p} - \hat{P}_{i+n_p}^{II}, \\
Error_p^{II} &= g_p\left(Error_p^{I}\right).
\end{aligned} \tag{11}$$

where $i$ is the current time $i$=1, 2, ..., 7, and with each $i$, $j$=0, 1, ..., 6. $\hat{P}_{i+n}$ is $n$ hours ahead predicted wind power, $n \in [6,24]$, $v$ is the wind speed observed in the turbine, $u$ represents the wind speed calculated from the mesoscale NWP wind model for the site.

Firstly, two large deep learning prediction models $f_p^I(.)$ and $f_p^{II}(.)$ for wind parks are created with only measured data and measured and NWP data, respectively, and the prediction error sequences are calculated separately. Then the relational function $g_p(.)$ of the two error sequences is established.

2. Student (small) training models for turbine $t$ and error functions transfer.

$$\hat{P}_{i+n_t\,Train}^{I} = f_t\left(P_{i-j_{t\,Train}}; v_{i-j_{t\,Train}}\right) \tag{12}$$

$$\hat{P}_{i+n_t\,Train}^{II} = \hat{P}_{i+n_t\,Train}^{I} + g_p(P_{i+n_t} - \hat{P}_{i+n_t\,Train}^{I})$$

Firstly, the learned large wind park prediction model $f_p^I(.)$ in the first stage is distilled to the small turbine model $f_t(.)$ with KD. Then the error relational function $g_p(.)$ for park is transferred to non-explicitly compensate for the lack of NWP information in the original prediction model $\hat{P}_{i+n_t\,Train}^{I}$ for turbines.

3. Turbine testing models

$$\hat{P}_{i+n_t\,Train}^{I} = \mathfrak{C}_e(\hat{P}_{i+n_t\,Train}^{II}) \tag{13}$$

$$\hat{P}_{i+n_t\,Test}^{I} = f_t\left(P_{i-j_{t\,Test}}; v_{i-j_{t\,Test}}\right) \tag{14}$$

$$\hat{P}_{i+n_t\,Test}^{II} = \mathfrak{C}^{-1}_e(\hat{P}_{i+n_t\,Test}^{I}) \tag{15}$$

Since the result labels in the test set should not be involved in testing, so the prediction errors in the test set cannot be engaged in the same way as the training for model correction. Therefore, primitive and inverse function transformation is introduced to link the original and improved model (with NWP information) for testing. Firstly, primitive function $\mathfrak{C}_e(.)$ is learned between the two predictions of the training set. Then the trained turbine model $f_t(.)$ is tested on the test set to get an original prediction $\hat{P}_{i+n_t\,Test}^{I}$. Finally, the improved test set model, delivering the final forecast $\hat{P}_{i+n_t\,Test}^{II}$, is obtained by the inverse operator $\mathfrak{C}^{-1}_e(.)$ on $\hat{P}_{i+n_t\,Test}^{I}$.

To summarize, the foregoing framework for wind turbine power prediction through KD, TL, and error correction is exhaustively illustrated in Fig. 4.

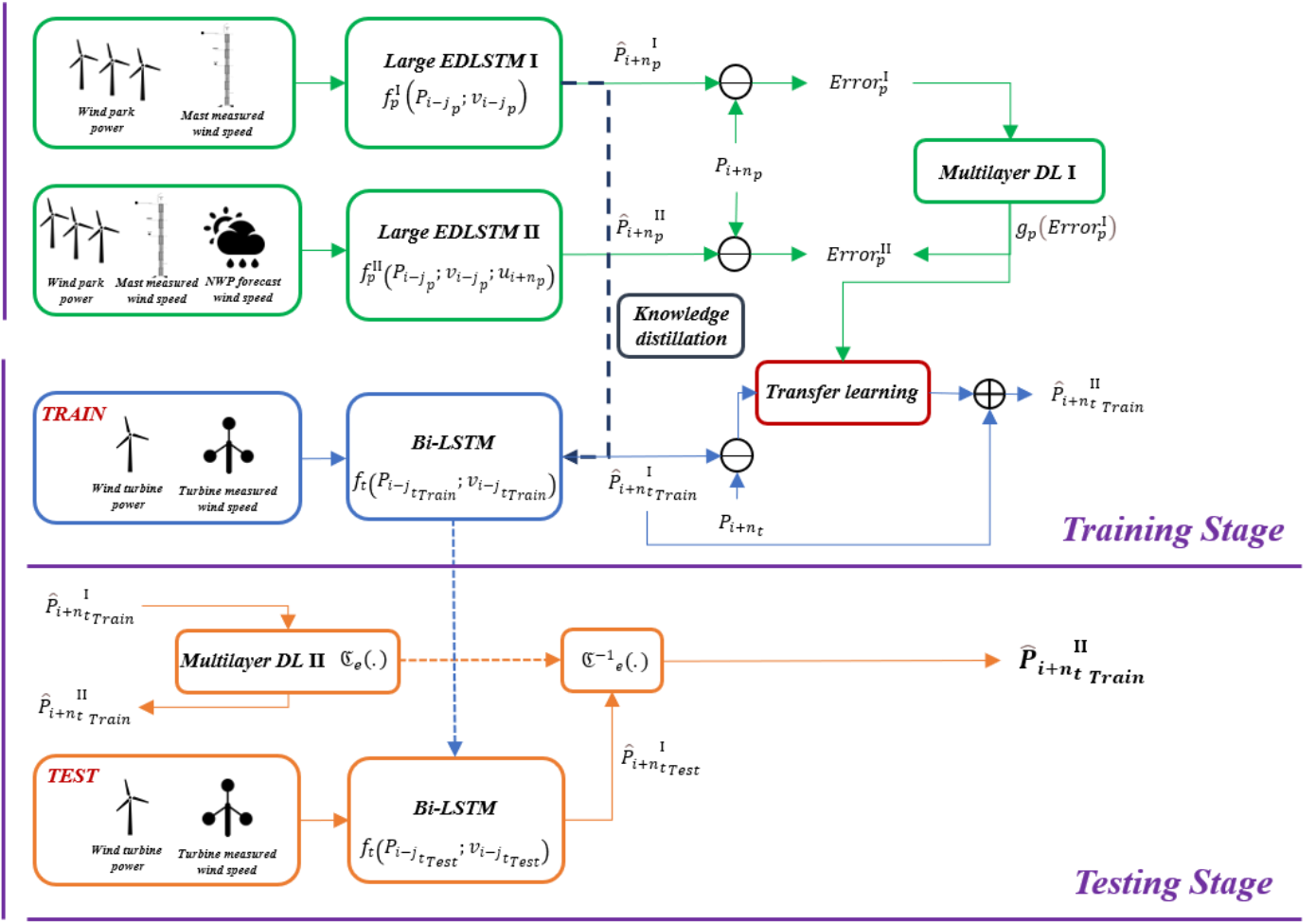

Fig. 4. The proposed KD structure for distilling knowledge from park to turbine.

## 4. Experiments

### 4.1. General experimental process

Firstly, the measured wind park and turbine 10-minute data are first interpolated into hourly data by averaging. And all data are standardized and divided into a training set, 65%, and a testing set, 35%. Then add white noise to expand the data to five times the original size (DA2 data augmentation in [17]). The training sets of the park (with 10-folds validations for improving training accuracy for its large forecasting model) and turbines are separately loaded into their corresponding algorithms to learn fine-tuned forecasting models and generate predicted power of the park and turbines. Subsequentially, the prediction errors can be calculated and transferred for the turbine prediction corrections if needed. Finally, the described turbine predictive modeling frames are tested with the turbines' testing sets and the performance is assessed and compared.

### 4.2. Performance evaluation

The data-driven wind power prediction is essentially categorized as a regression problem in machine learning, where the MSE controls the model's learning progress as the basic unit of the regressor loss function. Therefore, the Root Mean Square Error (RMSE) is the most dominant indicator to evaluate the models' performance. Besides, prediction Qualification Rate (QR) [47] is introduced to more physically evaluate models. RMSE emphasizes larger forecasting errors while the QR stresses smaller ones.

$$RMSE = \sqrt{\frac{\sum_{i=1}^{m}(P_i-\widehat{P}_i)^2}{m}} \tag{16}$$

$$QR = \frac{1}{m}\sum_{i=1}^{m} \begin{cases} 1, \left(1-\frac{|P_i-\widehat{P}_i|}{Cap}\right) \geqslant Q \\ 0, \left(1-\frac{|P_i-\widehat{P}_i|}{Cap}\right) < Q \end{cases} \tag{17}$$

where $P_i$ and $\widehat{P}_i$ represent standardized measured and corresponding predicted wind power, m is the sample number of the testing set. *Cap* donates the designed turbine capacity. *Q* is the quantile percentage of qualified predictions, selected as 90% in the work as QR90.

Moreover, as a crucial strength of the proposed model is merely training the large EDLSTM network on wind park data and transferring the network's knowledge to the small network for turbine prediction via the developed KD technique, the complexity and computing load of the proposed model could be anticipated to be considerably reduced. So, indices like number of model parameters, modeling time consumption and inference time on the edge are also compared with the rate between forecasting models.

Furthermore, since multiple groups of metrics are involved in the comparisons, a multivariate statistical test, the Friedman test [48], is applied to determine statistical significance in the metrics.

$H_o$: The column data fail to show a significant difference.

$H_a$: They display a significant difference.

The statistic *F* is in (18):

$$F = \frac{12t}{k(k+1)}\left[\sum_{i=1}^{k} r_i^2 - \frac{k(k+1)^2}{4}\right] \tag{18}$$

where *t* and *k* represent the number of samples and their columns, $r_i$ is the mean of row *i*, which follows $\chi^2_{(k-1)}$ distribution under $H_o$.

## 5. Results and discussion

This section presents the experimental results from a holistic range in three perspectives. Firstly, given that the KD and TL models serve as core pieces of the proposed method, and the foremost KD parameter $\alpha$ and TL parameter $\gamma$ remarkably affect the whole forecasting performance, it is necessary to find the most appropriate parameters first. The second part is the main content of the Section, which synthetically evaluates the proposed model by comparing it with the benchmark models. Moreover, a physical analysis of the predicted power is briefly undertaken.

### 5.1. Optimal TL and KD parameters finding

According to Eq. (7) and (8), the KD parameter $\alpha$ controls the weights of soft-loss from teacher model and hard-loss from student model. Therefore, an optimized KD architecture needs to be constructed before the implementation of the complete turbine prediction modeling. And based on to Eq. (4) the TL parameter $\gamma$ adjusts the contribution of MMD-based mechanism in TL process, with which the fine-tuned model will be more adaptable to new data.

A grid search is conducted for the parameter in the range of 0 to 1 in steps of 0.2, and the corresponding performance RMSE of the KD turbine prediction model in Fig. 2 (also Algorithm 1.1) with finding $\alpha$ is illustrated in Fig. 5 (a). The overall RMSE, for all targeted turbines, of KD is seen to exhibit nonuniform downward trends as $\alpha$ increases from 0 to 1. However, the RMSE variations and $\alpha$-values corresponding to the minimum points vary for different turbines. The KD models for the turbines, except for T3 (0.6) and T5 (1), are optimized for an $\alpha$ of 0.8. Intuitively, an interpretation could be that the student model learns more from the complex teacher when the $\alpha$ parameter is bigger, but completely abandoning the student's own loss function may also lead to underfitting on the training set. Therefore, $\alpha = 0.8$ is chosen as the optimized parameter of KD for all the five turbines in the following trials for $\gamma$ determinations.

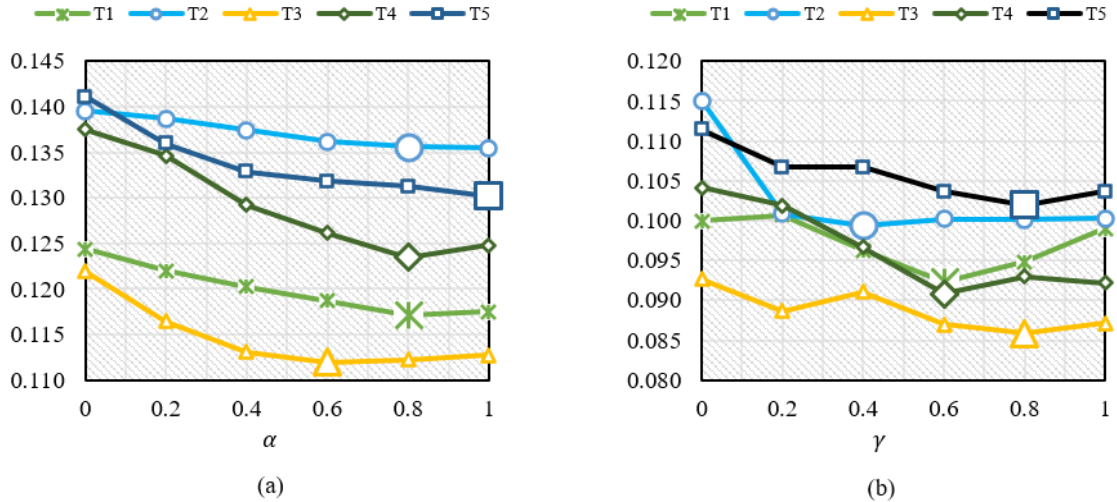

Fig. 5. The grid search for KD and TL parameters with indexing by RMSE (The points corresponding to the best performance is marked with bigger icons): (a) KD parameter $\alpha$ and TL parameter $\gamma$, (b) TL parameter $\gamma$.

Set the $\alpha$ of KD to 0.8 for all wind turbine prediction frames and then proceed to correct prediction errors with TL from the park to turbines. The same grid search is also made for determining a proper TL parameter $\gamma$ for error transferring. The performance comparison is displayed in Fig. 5 (b). The RMSE also shows a decrease with increasing $\gamma$ parameter, but not significantly, and even displays a decreasing and then increasing shape in T1 and T4. Besides, algorithmically, $\gamma$ controls the contribution of the comparison between wind field and turbine data in the TL, so its optimal value changes with the turbine data and TL internal states. Thus, it is hardly applicable to determine each turbine's TL with a single $\gamma$ parameter. For T1 and T4, $\gamma$ is 0.6; for T3 and T5, $\gamma$ is 0.8; $\gamma$ is 0.4 for T2.

Therefore, the two dominant parameters for the proposed turbine prediction framework are determined with the grid search and will be applied in the following experiments.

## 5.2. Multistep turbine power prediction

Two crucial parts are involved in the proposed model, i.e., KD and TL error correction (if the latter is involved, the Algorithm 1.3. must be included for the testing set) for turbine power predictions.

To explore the contribution of each part, several model variants are derived for the comprehensive ablation investigation. The models' description and their inputs are listed in Table 2.

Table 2. The general description for the used models.

| Model | Description | Inputs (standardized data) |
| --- | --- | --- |
| BiLSTM | Simple Bi-LSTM for turbine prediction. | Turbine measured data, NWP for park |
| EDLSTM | [17] proposed EDLSTM framework for turbine prediction. | Turbine measured data, NWP for park |
| KD | Only Fig. 3 KD system without TL error correction | Park and Turbine measured data |
| EDED-TL | EDLSTM for both park and turbine forecasting and they are connected with TL error correction, no KD is involved. | Park and Turbine measured data, NWP for park |
| EDBi-TL | Without KD, EDLSTM for park forecasting, and Bi-LSTM for turbine and they are linked with TL error correction. | Park and Turbine measured data, NWP for park |
| KD-TL | The proposed predictive framework in which KD combines EDLSTM for park and Bi-LSTM for turbine, their relationship is found in forecasting by TL error correction. | Park and Turbine measured data, NWP for park |

The six models and corresponding inputs from the above table are individually applied to five selected wind turbines to predict power from 6 to 12 hours in advance. All prediction models yielded relatively satisfactory outputs within forecasting steps, all with RMSE < 0.15, which fully illustrates the powerful nonlinear mapping capability of the deep LSTM neural network-based models. The RMSE for different models with varying prediction steps is illustrated in Fig. 6.

Generally, all prediction models RMSE rises with increasing steps, but the rising rate is gradually slowing down. In virtually multiple steps, EDED-TL and proposed KD-TL perform best among all turbines' predictions.

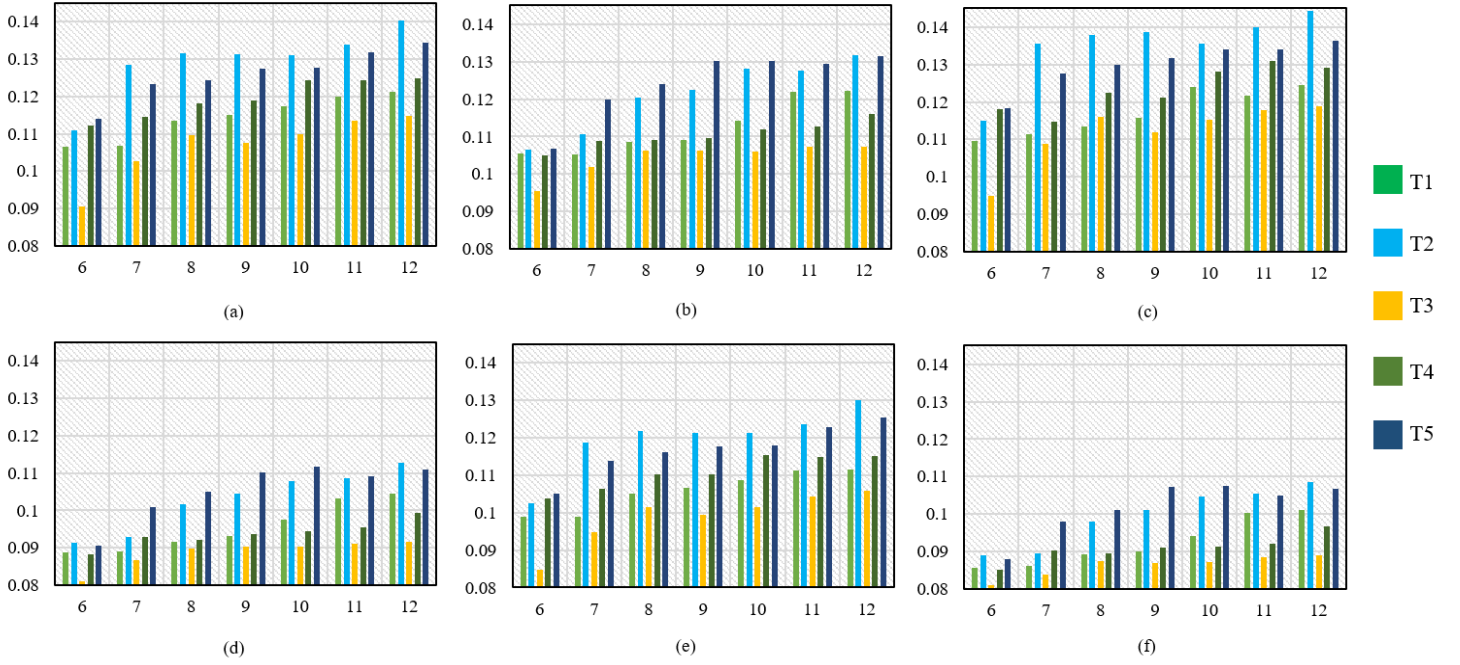

Fig. 6. The multistep RMSE of models with varying prediction algorithms for each turbine: (a) BiLSTM, (b) EDLSTM, (c) KD, (d) EDED-TL, (e) EDBi-TL, (f) KD-TL.

To quantitatively analyze the variations in multistep performance, the RMSE inter-step rising rates are calculated and the summarized geometric means, also with the Friedman tests, and STDs of these rates are taken separately for both turbines and algorithms are shown in the Table 3. Table 3(a) says the $p$-value of the Friedman statistic is much less than 0.05, indicating that the RMSE increase (from around 1.8% to 3.8%) with the step of the six prediction algorithms is significantly different. This also verifies the dominant influence of topography on turbine predictions. As can be observed from Table 3(b), its Friedman statistic gives a $p$-value of 0.9047 which means the RMSE inter-step increase is statistically identical for the models, which, to some extent, reveals that LSTM-based models can deeply and robustly mine the essential characteristics of the wind data and when a model has an initial edge, the edge carries over.

Table 3(a). The statistics for RMSE inter-step rising rates geomeans for different turbines.

| Rising rate between turbines $F = 16.4$; $p = 0.0025$ | T1 | T2 | T3 | T4 | T5 |
|---|---|---|---|---|---|
| **GeoMean** | 2.395% | 3.839% | 2.749% | 1.818% | 3.109% |
| **STD** | 0.0030 | 0.0032 | 0.0105 | 0.0020 | 0.0043 |

Table 3(b). The statistics for RMSE inter-step rising rates geomeans for various models.

| Rising rate with steps $F = 1.57$; $p = 0.9047$ | BiLSTM | EDLSTM | KD | EDED-TL | EDBi-TL | KD-TL |
|---|---|---|---|---|---|---|
| **GeoMean** | 2.853% | 2.576% | 2.676% | 2.716% | 2.810% | 2.567% |
| **STD** | 0.0096 | 0.0082 | 0.0103 | 0.0068 | 0.0097 | 0.0070 |

## 5.3. Holistic validity of the proposed model

Furthermore, the collective performance of various algorithms and different turbines are summarily compared in Fig. 7 with geometric means (including STD) in terms of algorithm and turbine perspectives. Fig. 7 (a) clearly presents the average performance for diverse algorithms in the multistep predictions. KD, despite distilling park's knowledge to turbine, performs the worst, which is because the weather forecasts are not included in the model's inputs, thus the prediction accuracy on historical measured data only is unsatisfactory even with advanced methods. Two baselines with NWP inputs, BiLSTM and EDLSTM, the latter is obviously better than the former, are slightly inferior to EDBi-TL. This indicates that turbine prediction is improved by TL of error correction even without the KD mechanism. And the improvement is considerably bigger than directly incorporating the wind park's NWP data into the turbine predictions. By discovering the advantage of EDED-TL over EDBi-TL, it means that the simple Bi-LSTM, compared to EDLSTM, may not learn the deep features of the turbine data well. Noteworthy, the proposed KD-TL delivers the lowest RMSE, (averagely 21.14%, 7.89%, 23.92%, 3.26%, and 14.86% lower than BiLSTM, EDLSTM, KD, EDED-TL, and EDBi-TL.) which proves the superiority of the proposed models among its competitors.

Fig. 7 (b) summarizes the prediction performance of the algorithms for varying turbines. The RMSE for turbines in flat terrains, such as T1 Plateau and T3 Lakeside, is minimal within an STD. By contrast, the distinctive fjord landscape of the Norwegian coast makes T2 Valley, T4 Hilltop, and T5 Seaside somewhat challenging. But the proposed framework holistically addresses these challenges and provides competitive predictions in these complex terrains (the KD-TL RMSE of these three turbines is smaller than RMSE of some baselines for turbines on flat ground.).

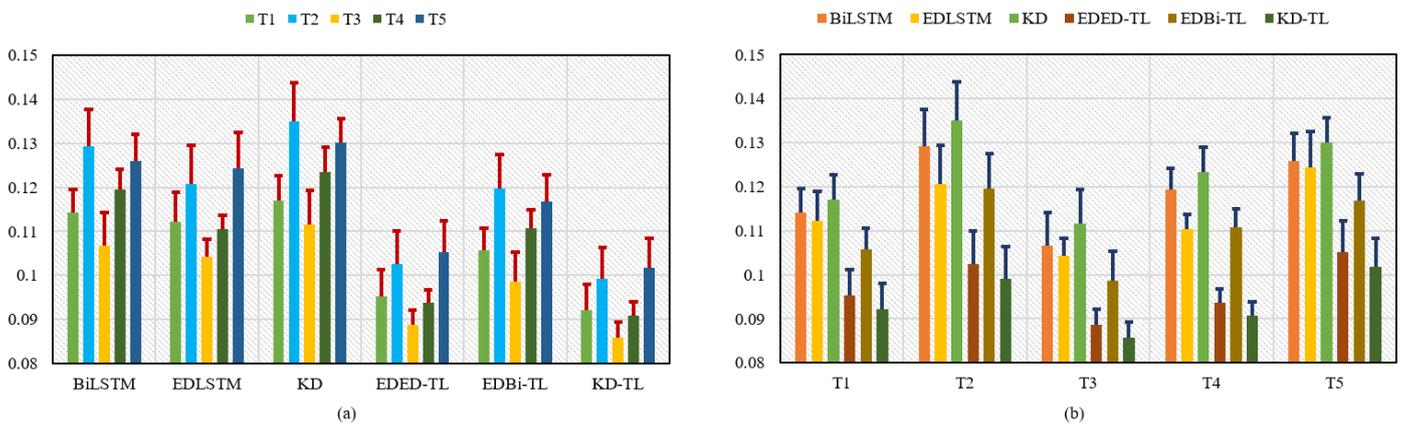

Fig. 7. The overall average RMSE of multistep prediction models with STD: (a) Predictive algorithm perspective, (b) Different turbine perspective.

Finally, the proposed KD-TL has the best performance, even lower than EDED-TL does, both in mentioned two perspectives, which demonstrates that the learned turbine Bi-LSTM (student) with KD from park EDLSTM is even slightly better than the complex EDLSTM, consuming much computing resources and time, for turbine itself. This suggests certain overfitting in the large

EDLSTM on the turbine smaller, compared with park, data. Moreover, its advantage over the EDED-TL also exists in model's complexity. The number of EDED-TL parameters is around 3.8 times bigger than the number of KD-TL (also means around 20 times for five turbines' forecasts.). And the former also spends around 2.3 folds as much time as the latter does on edge inference, which meets excepted diminished complexity and rapidity for the proposed model in engineering operations.

To further delve into the turbine power predictions by different. This subsection briefly explores the physical interpretation of the outcomes. Fig. 8 exhibits the geometric QR90 average of five turbines, from 6 to 12 steps, an increase of proposed KD-TL in comparison to other prediction models. As this metric focuses more on predictions within ten percent of the designed power difference between the predicted and real values, a high QR90 reflects that the prediction method gives a more trustworthy result. It is seen that compared to BiLSTM, EDLSTM, KD, EDED-TL, and EDBi-TL, KD-TL offers improvements of QR90 over 14.07%, 11.35%, 14.96%, 1.07%, and 7.86% within one STD respectively.

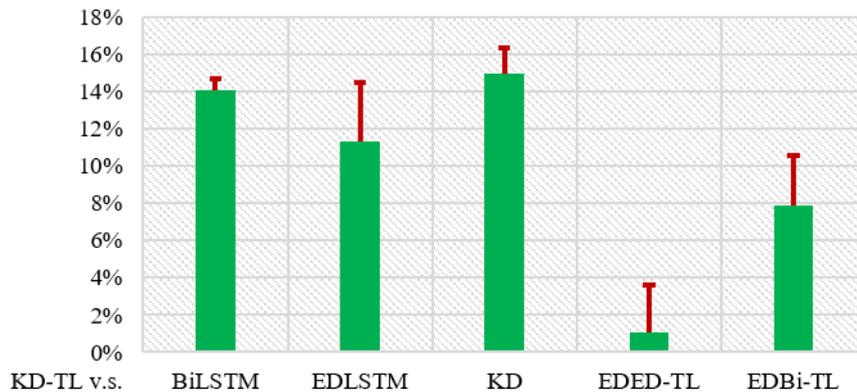

Fig. 8. The overall average QR90 increase rate of the proposed multistep KD-TL prediction versus other models.

In summary, the proposed model enables more precise and economic multistep turbine power prediction than the baseline models and such advantages exist for turbines on different topographic conditions.

## 6. Conclusions

This paper exploits a deep learning wind power prediction framework. It bridges large (park with big data) and small-scale (turbine with small data) forecasting through a proposed knowledge distillation regression approach, and maps park-scale weather forecasts non-explicitly to turbine-scale by TL-based prediction error corrections. The following conclusions are drawn from experiments on turbine predictions under five types of topography in a park in the Arctic.

As the turbine prediction models get knowledge by KD from the park model, it is vital to optimizing both contributions in the final prediction model (loss function in deep learning). The contributions' parameters for the KD are found as 0.8. And the parameters for error corrections change with turbines and their scope is determined to be 0.4 to 1.

The proposed KD-TL multistep power prediction framework is extraordinarily effective and robust by leveraging the data and reinforcing the nonlinear capabilities of the model based on the experimental comparisons. Compared to its competitors, the overall average effectiveness, by RMSE, is respectively improved from high to low: 23.9%, 21.1%, 14.9%, 7.9%, and 3.3%. The effectiveness is also verified with a metric with physical meaning, QR90. Moreover, KD-TL, thanks to its adequate utilization of weather forecast information, yields satisfactory outcomes in predicting wind turbines in complex terrain, normally challenging, as well. Finally, the complexity and response time of KD-TL decreases multiplicatively over its closest competitor, which enables the proposed approach to have extensive strengths in engineering deployment.

Furthermore, consequent policy recommendations could be inferred:

The assembled historical data should be sufficiently exploited to build accurate large models applicable to the local natural conditions in regional energy industries. With know, a smaller, easy-to-deploy model could be developed in practical applications by edge computing methods. In data-oriented engineering, initial model errors, usually ignored or dismissed as unfavorable, are valuable that can be extracted by progressively state-of-the-art deep learning. The extracted information may help the subsequent modeling through appropriate transfer methodology.

## Acknowledgments

The open-access publication charges for this article have been funded by a grant from the publication fund of UiT The Arctic University of Norway. And thanks to the support of UiT Arctic Centre for Sustainable Energy.

## Competing interests

The authors declare that they have no known competing financial interests or personal relationships that could have appeared to influence the work reported in this paper.

## Data availability

The NWP data is publicly available from The Norwegian Meteorological Institute. The measured wind data from Fakken wind park is the property of the power company Troms Kraft AS. These data are available on reasonable request from the authors.

## References

[1] S. Ahmed, *Wind energy: theory and practice*. PHI Learning Pvt. Ltd., 2015.
[2] W. C. Radünz *et al.*, "The variability of wind resources in complex terrain and its relationship with atmospheric stability," *Energy Conversion and Management,* vol. 222, p. 113249, 2020.

[3]     S. K. Salman and A. L. Teo, "Windmill modeling consideration and factors influencing the stability of a grid-connected wind power-based embedded generator," *IEEE Transactions on Power Systems,* vol. 18, no. 2, pp. 793-802, 2003.

[4]     H. Liu and C. Chen, "Data processing strategies in wind energy forecasting models and applications: A comprehensive review," *Applied Energy,* vol. 249, pp. 392-408, 2019.

[5]     H. Liu, C. Chen, X. Lv, X. Wu, and M. Liu, "Deterministic wind energy forecasting: A review of intelligent predictors and auxiliary methods," *Energy Conversion and Management,* vol. 195, pp. 328-345, 2019.

[6]     Z. Qian, Y. Pei, H. Zareipour, and N. Chen, "A review and discussion of decomposition-based hybrid models for wind energy forecasting applications," *Applied energy,* vol. 235, pp. 939-953, 2019.

[7]     S. Al-Yahyai, Y. Charabi, and A. Gastli, "Review of the use of numerical weather prediction (NWP) models for wind energy assessment," *Renewable and Sustainable Energy Reviews,* vol. 14, no. 9, pp. 3192-3198, 2010.

[8]     J. B. Olson *et al.*, "Improving wind energy forecasting through numerical weather prediction model development," *Bulletin of the American Meteorological Society,* vol. 100, no. 11, pp. 2201-2220, 2019.

[9]     J. González-Sopeña, V. Pakrashi, and B. Ghosh, "An overview of performance evaluation metrics for short-term statistical wind power forecasting," *Renewable and Sustainable Energy Reviews,* vol. 138, p. 110515, 2021.

[10]    A. Dupré, P. Drobinski, B. Alonzo, J. Badosa, C. Briard, and R. Plougonven, "Sub-hourly forecasting of wind speed and wind energy," *Renewable Energy,* vol. 145, pp. 2373-2379, 2020.

[11]    H. Wang, Z. Lei, X. Zhang, B. Zhou, and J. Peng, "A review of deep learning for renewable energy forecasting," *Energy Conversion and Management,* vol. 198, p. 111799, 2019.

[12]    X. Deng, H. Shao, C. Hu, D. Jiang, and Y. Jiang, "Wind power forecasting methods based on deep learning: A survey," *Computer Modeling in Engineering and Sciences,* vol. 122, no. 1, p. 273, 2020.

[13]    Y. Wang, R. Zou, F. Liu, L. Zhang, and Q. Liu, "A review of wind speed and wind power forecasting with deep neural networks," *Applied Energy,* vol. 304, p. 117766, 2021.

[14]    G. Giebel and G. Kariniotakis, "Wind power forecasting—A review of the state of the art," *Renewable energy forecasting,* pp. 59-109, 2017.

[15]    L. Lazić, G. Pejanović, M. Živković, and L. Ilić, "Improved wind forecasts for wind power generation using the Eta model and MOS (Model Output Statistics) method," *Energy,* vol. 73, pp. 567-574, 2014.

[16]    W. Y. Cheng, Y. Liu, A. J. Bourgeois, Y. Wu, and S. E. Haupt, "Short-term wind forecast of a data assimilation/weather forecasting system with wind turbine anemometer measurement assimilation," *Renewable Energy,* vol. 107, pp. 340-351, 2017.

[17]    H. Chen, Y. Birkelund, and Q. Zhang, "Data-augmented sequential deep learning for wind power forecasting," *Energy Conversion and Management,* vol. 248, p. 114790, 2021.

[18]    N. Chen, Z. Qian, I. T. Nabney, and X. Meng, "Wind power forecasts using Gaussian processes and numerical weather prediction," *IEEE Transactions on Power Systems,* vol. 29, no. 2, pp. 656-665, 2013.

[19]    Q. Yang, Y. Zhang, W. Dai, and S. J. Pan, *Transfer learning*. Cambridge University Press, 2020.

[20]  Q. Hu, R. Zhang, and Y. Zhou, "Transfer learning for short-term wind speed prediction with deep neural networks," *Renewable Energy,* vol. 85, pp. 83-95, 2016.

[21]  A. S. Qureshi, A. Khan, A. Zameer, and A. Usman, "Wind power prediction using deep neural network based meta regression and transfer learning," *Applied Soft Computing,* vol. 58, pp. 742-755, 2017.

[22]  H. Yin, Z. Ou, J. Fu, Y. Cai, S. Chen, and A. Meng, "A novel transfer learning approach for wind power prediction based on a serio-parallel deep learning architecture," *Energy,* vol. 234, p. 121271, 2021.

[23]  X. Liu, Z. Cao, and Z. Zhang, "Short-term predictions of multiple wind turbine power outputs based on deep neural networks with transfer learning," *Energy,* vol. 217, p. 119356, 2021.

[24]  B.-M. Hodge *et al.*, "Wind power forecasting error distributions: An international comparison," National Renewable Energy Lab.(NREL), Golden, CO (United States), 2012.

[25]  M. Ding, H. Zhou, H. Xie, M. Wu, Y. Nakanishi, and R. Yokoyama, "A gated recurrent unit neural networks based wind speed error correction model for short-term wind power forecasting," *Neurocomputing,* vol. 365, pp. 54-61, 2019.

[26]  Z. Sun and M. Zhao, "Short-term wind power forecasting based on VMD decomposition, ConvLSTM networks and error analysis," *IEEE Access,* vol. 8, pp. 134422-134434, 2020.

[27]  Z. Liang, J. Liang, C. Wang, X. Dong, and X. Miao, "Short-term wind power combined forecasting based on error forecast correction," *Energy Conversion and Management,* vol. 119, pp. 215-226, 2016.

[28]  I. Goodfellow, Y. Bengio, and A. Courville, *Deep learning*. MIT press, 2016.

[29]  G. Hinton, O. Vinyals, and J. Dean, "Distilling the knowledge in a neural network," *arXiv preprint arXiv:1503.02531,* vol. 2, no. 7, 2015.

[30]  Y. Wang, Y. Liu, L. Li, D. Infield, and S. Han, "Short-term wind power forecasting based on clustering pre-calculated CFD method," *Energies,* vol. 11, no. 4, p. 854, 2018.

[31]  L. Liu and Y. Liang, "Wind power forecast optimization by integration of CFD and Kalman filtering," *Energy Sources, Part A: Recovery, Utilization, and Environmental Effects,* vol. 43, no. 15, pp. 1880-1896, 2021.

[32]  X. Ying, "An overview of overfitting and its solutions," in *Journal of Physics: Conference Series*, 2019, vol. 1168, no. 2: IOP Publishing, p. 022022.

[33]  G. James, D. Witten, T. Hastie, and R. Tibshirani, *An introduction to statistical learning*. Springer, 2013.

[34]  C. Ren and Y. Xu, "Transfer learning-based power system online dynamic security assessment: Using one model to assess many unlearned faults," *IEEE Transactions on Power Systems,* vol. 35, no. 1, pp. 821-824, 2019.

[35]  L. Torrey and J. Shavlik, "Transfer learning," in *Handbook of research on machine learning applications and trends: algorithms, methods, and techniques*: IGI global, 2010, pp. 242-264.

[36]  C. Tan, F. Sun, T. Kong, W. Zhang, C. Yang, and C. Liu, "A survey on deep transfer learning," in *International conference on artificial neural networks*, 2018: Springer, pp. 270-279.

[37]  L. Wen, L. Gao, and X. Li, "A new deep transfer learning based on sparse auto-encoder for fault diagnosis," *IEEE Transactions on systems, man, and cybernetics: systems,* vol. 49, no. 1, pp. 136-144, 2017.

[38]  C. Buciluă, R. Caruana, and A. Niculescu-Mizil, "Model compression," in *Proceedings of the 12th ACM SIGKDD international conference on Knowledge discovery and data mining*, 2006, pp. 535-541.

[39]  J. Gou, B. Yu, S. J. Maybank, and D. Tao, "Knowledge distillation: A survey," *International Journal of Computer Vision,* vol. 129, no. 6, pp. 1789-1819, 2021.

[40]   L. Yuan, F. E. Tay, G. Li, T. Wang, and J. Feng, "Revisiting knowledge distillation via label smoothing regularization," in *Proceedings of the IEEE/CVF Conference on Computer Vision and Pattern Recognition*, 2020, pp. 3903-3911.

[41]  G. Chen, W. Choi, X. Yu, T. Han, and M. Chandraker, "Learning efficient object detection models with knowledge distillation," *Advances in neural information processing systems,* vol. 30, 2017.

[42]   M. R. U. Saputra, P. P. De Gusmao, Y. Almalioglu, A. Markham, and N. Trigoni, "Distilling knowledge from a deep pose regressor network," in *Proceedings of the IEEE/CVF International Conference on Computer Vision*, 2019, pp. 263-272.

[43]  Q. Xu, Z. Chen, M. Ragab, C. Wang, M. Wu, and X. Li, "Contrastive adversarial knowledge distillation for deep model compression in time-series regression tasks," *Neurocomputing,* 2021.

[44]  M. Kang and S. Kang, "Data-free knowledge distillation in neural networks for regression," *Expert Systems with Applications,* vol. 175, p. 114813, 2021.

[45]  J. B. Bremnes and G. Giebel, "Do regional weather models contribute to better wind power forecasts," *The Norwegian Meteorological Institute,* 2017.

[46]  H. Chen and R. Staupe-Delgado, "Exploiting more robust and efficacious deep learning techniques for modeling wind power with speed," *Energy Reports,* vol. 8, pp. 864-870, 2022.

[47]   L. Lijuan, L. Hongliang, W. Jun, and B. Hai, "A novel model for wind power forecasting based on Markov residual correction," in *IREC2015 The Sixth International Renewable Energy Congress*, 2015: IEEE, pp. 1-5.

[48]  J. D. Gibbons and S. Chakraborti, *Nonparametric Statistical Inference: Revised and Expanded*. CRC press, 2014.